\documentclass[10pt, a4paper]{article}

\usepackage[final]{lrec2026} 

\usepackage{graphicx}
\usepackage{tabularx,booktabs,array}
\newcolumntype{Y}{>{\centering\arraybackslash}X}
\renewcommand{\arraystretch}{1.1}
\usepackage{tipa}
\usepackage{tabularx,booktabs}

\title{English to Central Kurdish Speech Translation: Corpus Creation, Evaluation, and Orthographic Standardization}

\name{Mohammad Mohammadamini\textsuperscript{1}\quad
      Daban Q.~Jaff\textsuperscript{2,3}\quad
      Josep Crego\textsuperscript{4}\quad \\
      \textbf{\large Marie Tahon}\textsuperscript{1}\quad
      \textbf{\large Antoine Laurent}\textsuperscript{1}}

\address{\textsuperscript{1}\,LIUM, Le Mans University, Le Mans, France\\
         \textsuperscript{2}\,Erfurt University, Erfurt, Germany\\
         \textsuperscript{3}\,Koya University, Koysinjaq, Iraq\\
         \textsuperscript{4}\,SYSTRAN (ChapsVision), Paris, France\\[2pt]
         \texttt{\{mohammad.mohammadamini,marie.tahon,antoine.laurent\}@univ-lemans.fr} \\
         \texttt{daban.hamad\_ameen@uni-erfurt.de \quad jcrego@chapsvision.com}}

\abstract{
We present KUTED, a speech-to-text translation (S2TT) dataset for Central Kurdish, derived from TED and TEDx talks. The corpus comprises 91{,}000 sentence pairs, including 170 hours of English audio, 1.65 million English tokens, and 1.40 million Central Kurdish tokens. We evaluate KUTED on the S2TT task and find that orthographic variation significantly degrades Kurdish translation performance, producing nonstandard outputs. To address this, we propose a systematic text standardization approach that yields substantial performance gains and more consistent translations. On a test set separated from TED talks, a fine-tuned Seamless model achieves 15.18 BLEU, and we improve Seamless baseline by 3.0 BLEU on the FLEURS benchmark. We also train a Transformer model from scratch and evaluate a cascaded system that combines Seamless (ASR) with NLLB (MT).
\\ \newline 
}

\begin{document}

\maketitleabstract

\section{Introduction}
Speech translation maps source-language audio to target-language text or speech. By output modality, it is categorized as Speech-to-Text Translation (S2TT) \cite{s2tt} or Speech-to-Speech Translation (S2ST) \cite{s2st}. Early work typically uses a cascaded pipeline in which Automatic Speech Recognition (ASR) produces source-language text that Machine Translation (MT) then translates into the target text \cite{cascade}. Recent advances in Transformer-based architectures, both encoder–decoder models \cite{seamless} and large language model (LLM) approaches \cite{llms2tt}, broaden end-to-end S2TT and S2ST research, yet the field remains constrained by limited paired speech–text and speech–speech resources for most languages.

In the current study, we present a S2TT dataset for Central Kurdish (ISO 639: CKB). Kurdish (ISO 639: KUR) is an Indo-European language spoken by an estimated 36.4–45.6 million native speakers across Kurdistan (spanning Turkey, Iran, Iraq, and Syria) and in diaspora communities in Europe and North America. It comprises six dialects: Northern Kurdish (KMR), Central Kurdish (CKB), Southern Kurdish (SDH), Laki (LKI), Zaza (DIQ), and Hawrami (HAQ) \cite{jafar, hassanpour}. We focus on Central Kurdish (CKB), spoken by nearly 8 million native speakers \cite{jafar}, primarily written in a modified Arabic script and recognized as an official language in Iraq.

To address data scarcity for S2TT in Kurdish, we introduce KUTED (\emph{Kurdish TED}), a corpus derived from TED and TEDx talks. KUTED contains roughly 170 hours of English speech, transcribed in English and translated into Central Kurdish. We describe data collection, alignment, and cleaning; evaluate audio and text alignment and translation quality with human evaluators; and introduce a method for detecting misaligned audio files. Beyond limited data, Kurdish MT faces substantial orthographic variability \cite{jira}, which increases data sparsity and degrades translation quality. Therefore, we propose a systematic orthographic standardization approach intended to generalize to future Central Kurdish MT research and to other languages with similar challenges.

We evaluate KUTED in both end-to-end (E2E) and cascaded S2TT settings. For E2E, we fine-tune several Seamless models and analyze how alternative standardization methods affect system quality. Because text-to-text (T2TT) MT is typically better resourced than speech translation and our direction is from a high-resource language (English) to a low-resource one (Central Kurdish), we hypothesize that a strong ASR system followed by MT can yield superior results. Accordingly, we run cascaded experiments with a fine-tuned Seamless model for ASR and a fine-tuned NLLB model for MT, and we also train a Transformer-based S2TT system from scratch to evaluate KUTED without relying on pretrained models.

\section{Related Work}
Recent efforts to build speech translation datasets for high-resource languages have yielded several large-scale resources. Aug-LibriSpeech extends LibriSpeech with French translations, providing a 236-hour EN$\rightarrow$FR S2TT corpus \cite{librifrench}. CoVoST~2 is the largest publicly available speech translation corpus, offering two-way S2TT from English to 15 languages and from 21 languages to English \cite{covost, covost2}. VoxPopuli is a multi-way speech translation corpus constructed primarily from European Parliament event recordings and covers 15 European languages \cite{voxpopuli}. Three widely used datasets derive from TED: MUST-C (EN$\rightarrow$14 languages) \cite{must1, mustc}, TEDx (EN$\rightarrow$7 languages) \cite{tedx}, and Indic-TEDST (EN$\rightarrow$9 Indian languages) \cite{tedindic}. 

In a recent study \cite{kuvost}, Common Voice 18 was extended for En$\rightarrow$CKB S2TT. In this work, the English Common Voice transcriptions were translated into Central Kurdish, resulting in 1,003 hours of English speech paired with Kurdish translations. The evaluation shows high performance for in-domain speech translation; however, for out-of-domain translation on standard benchmarks such as FLEURS, the performance is limited. Another study provides 3,200 hours of CKB$\rightarrow$EN S2TT pseudo-labeled data using a pipeline composed of an ameliorated ASR and MT for Central Kurdish \cite{interspeech2025}. FLEURS is a multi-way S2ST/S2TT corpus spanning 101 languages which includes X$\rightarrow$CKB and CKB$\rightarrow$X \cite{FLEURS}. This dataset is designed for few-shot learning and widely used as a main speech translation evaluation benchmark. Table~\ref{datasets} summarizes the reviewed resources.

Data scarcity extends beyond speech translation to text-based MT. Although Kurdish appears in commercial systems such as Google Translate and Microsoft Bing, there is no reliable large-scale Kurdish parallel dataset available to the research community. Among MT datasets that include Kurdish, we note FLEURS \cite{FLEURS}, FLORES \cite{flores}, and the TICO-19 \cite{covid} benchmarks, which primarily serve for evaluation. In \cite{ahamdi2}, a web-scraped corpus presented that includes approximately 2.2k EN$\rightarrow$CKB pairs and about 12k KMR$\rightarrow$CKB pairs. Hawta is the largest parallel corpus developed for Central Kurdish, comprising roughly 300k EN$\rightarrow$CKB pairs, with a portion publicly available \cite{hawta}. The text-only version of the dataset introduced in this paper, can serve as valuable resource for Kurdish MT.

\begin{table*}[t]
\centering
\footnotesize
\begin{tabularx}{\textwidth}{@{} l Y Y Y Y @{}}
\toprule
\textbf{Dataset} & \textbf{Source} & \textbf{Target} & \textbf{Size (hours)} & \textbf{Kurdish} \\
\midrule
Must-C                   & EN                   & 14 langs             & [237,505] & $\times$ \\
TEDx                     & EN                   & 7 langs              & [15,189]  & $\times$ \\
Aug-LibriSpeech          & EN                   & FR                   & 212        & $\times$ \\
CoVoST~2 EN$\rightarrow$X & EN                   & 15 langs             & 415        & $\times$ \\
CoVoST~2 X$\rightarrow$EN & 21 langs             & EN                   & [3,225]   & $\times$ \\
VoxPopuli                & 15 European langs    & 15 European langs    & [1,463]   & $\times$ \\
Indic-TEDST              & EN                   & 9 Indian langs       & [2,100]   & $\times$ \\
FLEURS                   & 102 langs            & 102 langs            & 1,400 & $\surd$ \\
Kuvost                   & EN            & CKB            & 1,003 & $\surd$ \\
Pseudo-labeled                   & CKB            & EN            & 3,200 & $\surd$ \\

\bottomrule
\end{tabularx}
\caption{Speech translation datasets. The size column shows the range of speech data (hours) for language pairs in each dataset.}
\label{datasets}
\end{table*}

\section{Kurdish TED Speech Translation Corpus}
\label{sec:format}

\subsection{The Kurdish TED Translator Community}
Founded in 2015 by Kurdish translation enthusiasts, the Kurdish TED Translator Community grew rapidly at Koya University, where volunteers primarily from the Departments of Linguistics and English Literature translated a large share of the talks. Approximately 1{,}500 talks are translated through this initiative, with the remainder produced by other volunteers. We outline the workflow to help mobilize similar communities to create resources for speech translation and language technology more broadly. The end-to-end process—preparing students, translating, and publishing TED Talks—proceeds in three phases:

\begin{itemize}
    \item \textit{\textbf{Training phase:}} Senior students are invited to translate TED Talks. Training consists of four online workshops (three hours each over one week) covering: (1) basic translation techniques (e.g., treatment of proper and country names, figurative language), (2) Kurdish grammar, (3) punctuation conventions, and (4) hands-on guidance for using the Amara and CaptionHub\footnote{\url{https://www.captionhub.com/}} platforms that host TED transcripts.
    \item \textit{\textbf{Translation phase:}} Students claim assignments and apply the workshop guidelines. They work in groups of 4–5 to support one another with technical or linguistic challenges throughout the translation process.
    \item \textit{\textbf{Editing and feedback phase:}} An experienced translator (the instructor) reviews each submission, either flagging errors for revision or making minor edits and approving the translation for publication. In both cases, students are encouraged to compare their submissions with the final versions using Amara/CaptionHub tools to internalize corrections for future work.
\end{itemize}

\subsection{Data collection}
The corpus is derived from the TED and TEDx content. First, we collect all verified transcriptions with Kurdish translations from CaptionHub, yielding 2{,}133 caption files. The corresponding audio is obtained from the TED website and the TED and TEDx YouTube channels. Talks centered on music or performances are discarded. In total, we obtain 1{,}696 TED/TEDx talks with human-annotated captions. All audio is converted to 16 kHz, mono channel. We then categorize talks as \emph{noisy} or \emph{clean}; noisy talks—primarily from TEDx—contain background sounds (e.g., music, wind, animal noise). Of the 1{,}696 talks, 467 are noisy and 1{,}229 are clean.

\subsection{Text and speech alignment}
We perform text–audio alignment in four steps:
\begin{itemize}
    \item \textbf{TED-level audio realignment:} Many files begin with an introductory segment that shifts timings causing misalignment between audio and transcript. Because the offset varies, we manually inspect all files and remove the introduction to eliminate the mismatch.
    \item \textbf{Sub-sentence text realignment:}  We use timestamps from the English and Kurdish transcription files collected via CaptionHub. We then realign English–Kurdish pairs at the sub-sentence level. By sub-sentences, we mean transcriptions that are incomplete sentences, where a full sentence has been divided into several parts.  When the offset between corresponding sub-sentences is  less than 1 second, we realigned it directly. For larger offsets, we realign the sub-sentences bounded by $<SS1>$ and $<SS2>$ (aligned at time $T$) and by $<\!ES1\!>$ and $<\!ES2\!>$ (aligned at $T{+}E$), ensuring that no portion of the transcription within that span is misaligned.
    \item \textbf{Sentence-level text alignment:} We also align at the sentence level. Sentence boundaries in the English transcription are marked by “.”, “!”, and “?”.
    \item \textbf{Audio extraction:} For each aligned sentence, we take the start time of its first sub-sentence and the end time of its last sub-sentence as the boundaries of the corresponding audio segment.
\end{itemize}

\subsection{Data specifications}
Processing yields 91,000 audio segments. The total duration is 170 hours of speech, with 1.65 million English tokens and 1.40 million Central Kurdish space separated tokens. Further specifications are provided in Table \ref{tbl:specif}.

\begin{table}[t]
\centering
\footnotesize
\begin{tabularx}{0,9\columnwidth}{Xccccc}
\toprule
\textbf{Split} & \textbf{TEDs} & \textbf{Utt} &
\shortstack{\textbf{EN}\\\textbf{speech}} &
\shortstack{\textbf{EN}\\\textbf{tokens}} &
\shortstack{\textbf{CKB}\\\textbf{tokens}} \\
\midrule
Clean & 1,229 & 75k & 138h & 1.35m & 1.14m \\
Noisy &  467 & 16k &  32h & 0.30m & 0.26m \\
All   & 1,696 & 91k & 170h & 1.65m & 1.40m \\
\bottomrule
\end{tabularx}
\caption{KUTED corpus specifications}
\label{tbl:specif}
\end{table}

\begin{figure}[!ht]
\centering
\includegraphics[width=\columnwidth]{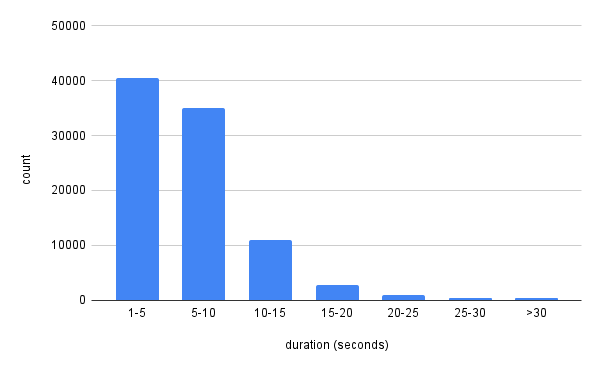}
\vspace{-0.2cm}
\caption{The duration distribution of speech files in KUTED dataset}
\label{fig:duration}
\end{figure}

The average duration of speech files is 6.73 seconds. The duration distribution of the audio files is shown in Figure \ref{fig:duration}.

There is a notable discrepancy between English and Kurdish token counts. Owing to its agglutinative, synthetic morphology, Kurdish often uses fewer tokens than English: while English frequently relies on separate words to express a concept, Kurdish fuses multiple grammatical elements into a single token to convey the same meaning \cite{hassanpour}. Consequently, English typically has more tokens than Kurdish within the same aligned parallel data. For example, the word \textipa{["ha\textbeltl m\"angIRtun]}, meaning “We have taken them,” corresponds to four tokens in English. We observe the same pattern in existing resources, including the Central Kurdish portions of FLORES \cite{flores} and FLEURS \cite{FLEURS}.

\subsection{Human data quality evaluation}\label{evaluation}
To ensure dataset quality, we conducted a multi-level human evaluation by sampling 1{,}000 pairs from the full corpus. The procedure was as follows:
\begin{itemize}
    \item \textbf{Audio alignment:} Listening to the 1{,}000 samples, we found that 922 (92.2\%) were correctly aligned. In 53 (5.3\%) samples, a few milliseconds at the beginning or end were missing or taken from adjacent segments—an offset negligible relative to the utterance length. In 16 (1.6\%) samples, the shift was more substantial (more than one word). We also identified 7 (0.7\%) misaligned samples and 2 (0.2\%) samples in which the audio was not English. Revisiting the source TED talks indicated these were occasional issues and that most other segments from the same talk were correctly aligned. In Section~\ref{misalign}, we propose an ASR-based method to detect and filter misaligned samples.
    \item \textbf{Text alignment:} We reviewed manually the selected samples for alignment between the English and Kurdish transcriptions. In all cases, text alignment was correct, as we had access to manually annotated and validated Kurdish translations.
    \item \textbf{Translation quality:} A subset of 500 pairs was assessed by three professional translators on four criteria:
    \begin{itemize}
        \item \textit{\textbf{Accuracy:}} Preservation of meaning and key lexical choices. 
        \item \textit{\textbf{Fluency:}} Grammaticality and naturalness.
        \item \textit{\textbf{Adaptation:}} Cultural appropriateness, including localization of idioms and metaphors.
        \item \textit{\textbf{Orthography:}} Conformity to Kurdish orthographic conventions.
    \end{itemize}
    
For each criterion, evaluators assigned discrete scores from 0 to 5 (0 = completely wrong; 1 = very low; 2 = low; 3 = average; 4 = good; 5 = very good). Table~\ref{tbl:evaluation} reports average scores per evaluator. Although there is a divergence of opinion among the three evaluators, the average rating scores for all metrics are above 4 out of 5. 
\end{itemize}

\begin{table}[!ht]
\centering
\footnotesize
\begin{tabularx}{\columnwidth}{@{} >{\raggedright\arraybackslash}X Y Y Y Y @{}}
\toprule
\textbf{Evaluator} & \textbf{Accuracy} & \textbf{Fluency} & \textbf{Adaptation} & \textbf{Orthog.} \\
\midrule
Evaluator~1 & 4.68 & 4.59 & 4.63 & 4.73 \\
Evaluator~2 & 4.30 & 4.22 & 4.16 & 4.41 \\
Evaluator~3 & 3.83 & 3.80 & 3.76 & 4.03 \\
\midrule
\textbf{Average} & \textbf{4.27} & \textbf{4.20} & \textbf{4.18} & \textbf{4.30} \\
\bottomrule
\end{tabularx}
\caption{Translation quality evaluation by professional translators.}
\label{tbl:evaluation}
\end{table}

\subsection{Audio misalignment detection using an ASR model}\label{misalign}
As noted in Section~\ref{evaluation}, a subset of audio segments is misaligned. We detect such cases by decoding all audio with a robust pretrained ASR system—specifically, the Seamless v2 Large model \cite{seamless}—and comparing the ASR hypothesis (H) to the reference transcript (R). We compute the normalized Levenshtein distance:
\begin{equation}
    D(R,H) = \frac{Lev(R,H)}{L(R+H)},
\end{equation}
where $Lev(R,H)$ is the Levenshtein distance \cite{jurafsky} between $R$ and $H$, and $L(R+H)$ is the length of the concatenation of $R$ and $H$. We mark a sample as misaligned if $D > 0.3$, a threshold determined empirically. Applying this filter removes approximately 4{,}000 samples. We categorize alignment errors as: (i) small or large boundary shifts at the beginning or end of the utterance, (ii) incorrect alignments, and (iii) non-English audio. Table~\ref{tbl:misalign} shows the distribution of these error types.
Using the proposed approach, we filtered out all incorrect and non-English samples. In addition, we reduced the number of shifted samples to a high degree.
\begin{table*}[t]
\centering
\footnotesize
\begin{tabularx}{\textwidth}{@{} >{\raggedright\arraybackslash}X Y Y Y Y Y Y @{}}
\toprule
 & \textbf{Samples} & \textbf{Correct} & \textbf{S-shift} & \textbf{L-shift} & \textbf{Incorrect} & \textbf{Non-EN} \\
\midrule
Before filtering & 1,000 & 925 & 53 & 16 & 7 & 2 \\
ASR filtering    & 1,000 & 966 & 30 &  4 & 0 & 0 \\
\bottomrule
\end{tabularx}
\caption{Audio misalignment detection. S-shift = small shift; L-shift = large shift; Incorrect = audio and transcription not aligned; Non-EN = audio not in English.}
\label{tbl:misalign}
\end{table*}
\subsection{Orthography standardization}\label{sec:standard}
The absence of a fully standardized writing system makes speech and text translation for low-resource languages more challenging. For Central Kurdish, orthographic variability directly affects benchmarking: systems may produce a correct token that is nevertheless scored as an error because its surface form does not match the reference. We address Central Kurdish orthographic standardization and its impact on speech translation performance. The main sources of variation are:

\begin{itemize}
    \item \textbf{Joined vs.\ non-joined words:} A major source of variability is whether words are written as single tokens or separated, especially with compound and derivational verbs. For instance, \textipa{[/bak\"aRhEn\"an/]} ‘use, utilize’ appears in at least four forms: \textipa{[/bak\"aRhEn\"an/]}, \textipa{[/ba k\"aR hEn\"an/]}, \textipa{[/bak\"aR hEn\"an/]}, and \textipa{[/ba k\"aRhEn\"an/]}. We unify such variants following guidance from the Kurdish Academy of Language \cite{academy}.
   
    \item \textbf{Loanwords and proper names:} Many loanwords and proper names admit multiple spellings (e.g., “culture,” “hydrogen”) \cite{jira}. We generally select the most frequent form attested in the corpus or the form recommended by the Kurdish Academy of Language \cite{academy}.
    
    \item \textbf{Inflectional and derivational affixes:} Several productive affix classes have multiple allomorphs. For example, the indefinite suffix appears in seven forms, though only two are recognized as standard by the Kurdish Academy of Language \cite{academy}. In this work, we standardize the allomorph sets listed in \cite{jira}.
\end{itemize}
In a systematic approach, we perform orthography standardization in three steps:

\textbf{N1) Normalization:} We use the AsoSoft text normalizer to apply Unicode correction, punctuation standardization, and number unification \cite{normalization}. Because Kurdish text may be typed with different keyboards and fonts, Unicode correction maps variant code points to a standardized Unicode form. Punctuation marks that serve the same function are normalized to a single convention, and numbers are unified in the Arabic-script form. In our experiments, we also separate punctuation from tokens, which substantially reduces the lexicon size.

\textbf{N2) General correction table:} We then apply a general correction table containing 19{,}700 pairs of misspelled and corrected tokens, derived from the most frequent words in the AsoSoft text corpus—the largest available Kurdish text corpus \cite{asosoftcorpus}. Approximately 7{,}700 exact matches from this table are found in our dataset and replaced throughout the corpus.

\textbf{N3) KUTED-specific correction table:} To obtain more reliable, standardized Kurdish transcriptions, we extract the unique token list from the normalized KUTED dataset and review all types occurring more than once. In total, 56{,}000 unique tokens are revised, yielding a new correction table with 11{,}860 misspelled/corrected pairs. Applying this table replaces about 150{,}000 token instances—roughly 10\% of all Kurdish tokens in the corpus. We release this correction table with the dataset to facilitate standardization in future work. The number of unique tokens at each step is shown in Table~\ref{tab1:normaluniq}. After the three steps of standardization, the number of tokens was reduced by half, which shows the significant impact of non-standardization on translation quality.

\begin{table}[!ht]
\centering
\footnotesize
\begin{tabularx}{\columnwidth}{l r r r r}
\toprule
\textbf{Normalization} & N0 & N1 & N2 & N3 \\
\midrule
\textbf{Unique Tokens} & 235,674 & 145,685 & 125,800 & 118,643 \\
\bottomrule
\end{tabularx}
\caption{Unique tokens after each step of normalization. N0 represents the original Kurdish transcriptions; N1–N3 are the three steps of normalization/standardization discussed in this section.}
\label{tab1:normaluniq}
\end{table}

\section{Translation systems}
\subsection{E2E S2TT}
\begin{figure}
\centering
\includegraphics[angle=0,origin=c,width=0.95\columnwidth,keepaspectratio]{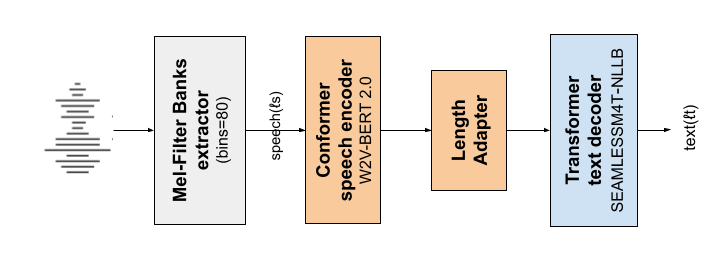}
\vspace{-0.1cm}
\caption{The Seamless V2 S2TT architecture}\label{fig:sl}
\end{figure}

The first E2E S2TT system used to evaluate the KUTED dataset is Seamless model. The architecture of Seamless model is shown in Figure \ref{fig:sl}. In this system, the speech encoder is W2V-BERT~2.0, which takes log-Mel filterbanks as input. The encoder is pretrained on 4.5 million hours of publicly available speech from 143 languages. A length adapter follows the encoder to align speech features with the text sequence, and the pretrained NLLB text decoder serves as the final component. The length adapter is a transformer module shrinks the speech representation.

In the Seamless model, these pretrained modules are fine-tuned for ASR and S2TT using data from 102 languages. While the documentation does not explicitly state whether Central Kurdish is included in W2V-BERT~2.0’s unsupervised pretraining, it is included in the NLLB pretrained text decoder. During end-to-end training of the full S2TT pipeline, EN$\rightarrow$CKB is also included. This stage uses 2{,}000 hours of pseudo-labeled data generated by a machine translation system. Further architectural and training details are provided in \cite{seamless}. We treat the pretrained components as the baseline and fine-tune them end-to-end on KUTED, aiming to demonstrate KUTED’s effectiveness for improving pretrained S2TT models.

\subsection{Cascaded S2TT}
\begin{figure}[!ht]
\centering
\includegraphics[origin=c,width=0.99\columnwidth,keepaspectratio]{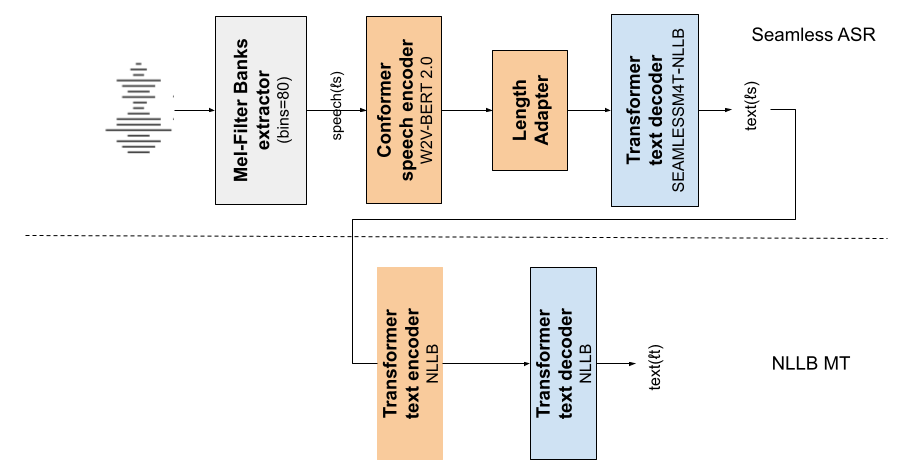}
\vspace{-0.1cm}
\caption{The cascade S2TT system}\label{fig:cascade}
\end{figure}

The cascaded speech translation system consists of two components in sequence: ASR followed by MT. For ASR, we use the Seamless model with the same architecture described for the end-to-end S2TT setup above. The ASR output is then passed to an NLLB 1.3B model, a dense encoder–decoder Transformer \cite{NLLB}. An overview of the cascaded S2TT pipeline is shown in Figure~\ref{fig:cascade}.

First, we fine-tune the pretrained Seamless ASR model on KUTED. The fine-tuned ASR then converts the English test speech to text. Next, we translate these ASR transcripts from English into Kurdish using an NLLB 1.3B model fine-tuned on the training portion of the aligned EN$\rightarrow$CKB text in KUTED. This cascaded experiment enables separate assessment of audio alignment (via ASR) and text alignment/translation (via MT).

\subsection{E2E S2TT from scratch}
While improving pretrained models such as Seamless demonstrates KUTED’s value, training an end-to-end (E2E) S2TT system from scratch provides a more rigorous test of corpus quality. Our third system is a Fairseq speech translation model trained solely on KUTED. The model comprises a Transformer-based speech encoder and a Transformer text decoder. In this setup, we do not use self-supervised pretrained S2TT components: the Fairseq speech encoder is pretrained only for English ASR (to avoid overfitting on a relatively small dataset), but the text decoder is not pretrained. We use a medium-size Transformer architecture with 76 million parameters \cite{fairseq}.

\section{Evaluation protocol}
To establish a fixed benchmark, we partition the data into train/validation/test splits. Of the 1{,}696 talks, we reserve 12 complete talks for validation and 16 complete talks for testing. For both validation and test, we ensure diversity in gender, age, and environmental noise. Table~\ref{division} reports the details of these splits.

\begin{table}[!ht]
\centering
\footnotesize
\begin{tabularx}{\columnwidth}{lYYY}
\toprule
\textbf{Metric} & \textbf{Train} & \textbf{Test} & \textbf{Validation} \\
\midrule
TEDs        & 1,668  & 16   & 12  \\
Men         & --    & 6    & 4   \\
Women       & --    & 6    & 4   \\
Children    & --    & 2    & 2   \\
Noisy TEDs  & 463   & 2    & 2   \\
Utterances  & 89,398 & 1,006 & 678 \\
\bottomrule
\end{tabularx}
\caption{Data partition for training, testing, and validation.}
\label{division}
\end{table}

In the test set, there are 6 complete talks by men, 6 by women, 2 featuring child speakers, and 2 categorized as noisy. As a secondary evaluation protocol, we also use FLEURS \cite{FLEURS} to demonstrate KUTED’s utility for out-of-domain speech translation. In our experiments, the KUTED test set serves as the primary benchmark, while FLEURS assesses generalizability to unseen domains.

\section{Experiments and results}
\subsection{Seamless E2E S2TT results}
The hyperparameters used to finetune the Seamless model are listed in the Table \ref{parameters}.

\begin{table}[!ht]
\centering
\footnotesize
\begin{tabularx}{0,5\columnwidth}{l r}
\toprule
\textbf{Parameter} & \textbf{Value} \\
\midrule
Learning rate & 1e-4 \\
Warmup steps  & 100  \\
Patience      & 50   \\
Batch size    & 6    \\
Max epochs    & 10   \\
\bottomrule
\end{tabularx}
\caption{Seamless fine-tuning hyperparameters}
\label{parameters}
\end{table}

Patience denotes the number of consecutive validation checks without improvement tolerated before early stopping halts training. The model is trained on three NVIDIA A100 GPUs. Utterances longer than 35\,seconds—which are very few in number (see Figure~\ref{fig:duration})—are excluded from training. Table~\ref{tab:results-e2e} reports results on the KUTED and FLEURS benchmarks.

\begin{table}[t]
\centering
\footnotesize
\setlength{\tabcolsep}{3pt}
\renewcommand{\arraystretch}{1.1}
\begin{tabularx}{0,75\columnwidth}{lcc}
\toprule
\textbf{System} & \textbf{KUTED} & \textbf{FLEURS} \\
\midrule
Seamless Baseline & 5.04 & 9.36 \\
Seamless KUTED    & 13.51 & 12.50 \\
\bottomrule
\end{tabularx}
\caption{E2E S2TT results for Seamless model on KUTED and FLEURS benchmarks.}
\label{tab:results-e2e}
\end{table}

The \textit{Seamless baseline} reports scores obtained with the pretrained Seamless model before fine-tuning, while \textit{Seamless KUTED} reports scores after fine-tuning on KUTED. The baseline achieves 5.04 BLEU on the KUTED (TED) benchmark and 9.36 BLEU on FLEURS. For both datasets, the corpus-independent N1 and N2 normalizations are applied. With standardized data and fine-tuning on KUTED, Seamless reaches 13.51 BLEU on KUTED and improves from 9.36 to 12.50 BLEU on FLEURS. These sizable in-domain and out-of-domain gains underscore KUTED’s impact on Central Kurdish S2TT.

\subsection{Impact of standardization on S2TT performance}
To quantify the effect of text standardization on speech translation, we run ablations over the normalization/standardization pipeline (N1–N3) described in Section~\ref{sec:standard}. Results are reported in Table~\ref{standardresults}.

\begin{table}[!ht]
\centering
\footnotesize
\begin{tabularx}{\columnwidth}{lYYY}
\toprule
\textbf{System} & \textbf{N1} & \textbf{N2} & \textbf{N3} \\
\midrule
Seamless Baseline & 4.71 & 5.04 & 5.47 \\
Seamless KUTED      & 11.34 & 13.51 & {15.18} \\
\bottomrule
\end{tabularx}
\caption{Impact of standardization on CKB S2TT. The results are BLEU score evaluated on KUTED test set.}
\label{standardresults}
\end{table}

The $N1$ condition applies general text normalization using the AsoSoft normalizer\footnote{\url{https://github.com/AsoSoft/AsoSoft-Library}}. In $N2$, we apply a general correction table, and $N3$ standardizes Kurdish using a KUTED-specific correction table derived from the corpus’s unique token list (see Section~\ref{sec:standard}).

As the results show, the Seamless baseline exhibits little to no change across the normalization steps—unsurprising, since it is not trained on standardized Kurdish. By contrast, the fine-tuned Seamless model on KUTED improves substantially, with BLEU rising from 11.34 to 15.18. This supports the view that weak BLEU scores in Kurdish MT/S2TT are driven in part by orthographic variation. After $N3$, the trained system appears to internalize most of the standardized Kurdish orthographic rules applied during training.

\subsection{Seamless–NLLB cascade system results}
In the cascade setup, Seamless ASR fine-tuning hyperparameters match Table~\ref{parameters}. The NLLB-1.3B model is fine-tuned with a learning rate of $1\times10^{-4}$ using Adam for $100$k iterations. Results for the cascade system are reported in Table~\ref{tab1:cascade}.

\begin{table*}[t]
\centering
\footnotesize
\begin{tabularx}{\textwidth}{lYY}
\toprule
\textbf{System} & \textbf{ASR Seamless (↓WER)} & \textbf{MT NLLB (↑BLEU)} \\
\midrule
Seamless–NLLB Baseline & 21.62 &  9.25 \\
Seamless–NLLB KUTED   &{8.62} &{15.57} \\
\bottomrule
\end{tabularx}
\caption{Cascade S2TT system results (lower WER is better; higher BLEU is better).}
\label{tab1:cascade}
\end{table*}

The first rows report the baseline system, in which we evaluate ASR and MT without any fine-tuning on KUTED. On the KUTED test set, the Seamless v2 Large ASR attains a WER of 21.0. The resulting transcripts are then translated directly by the NLLB 1.3B model, yielding 9.25 BLEU. When we fine-tune NLLB on the KUTED training split, it translates the ASR output to 15.57 BLEU—approximately matching the end-to-end S2TT result. Moreover, the lower WER achieved by ASR after fine-tuning on KUTED further corroborates the high-quality alignment in KUTED and its suitability for English ASR.

\subsection{Fairseq E2E S2TT results}
The preceding experiments show that KUTED effectively improves pretrained models that support Kurdish. To assess corpus quality independent of such pretraining, we also train a model from scratch. Specifically, we train a Fairseq Transformer with 76\,M parameters. Following \cite{fairseq}, the speech encoder is pretrained for English ASR to mitigate overfitting, while the text decoder is not pretrained. Training uses a learning rate of $2\times 10^{-3}$ with the Adam optimizer for $20$k iterations on three RTX~8000 GPUs, with an aggregated batch size of 96. We use a BPE tokenizer \cite{bpr} with a vocabulary of 10{,}000. Averaging the last $10$ checkpoints yields 7.90 BLEU on the KUTED test set. The relatively low scores given by this model can be attributed to several factors, such as the lack of SSL, and the limited size of KUTED. However, it is clear that the KUTED corpus alone cannot achieve satisfactory results and should be used alongside other corpora for training EN→CKB S2TT models.

\subsection{NLLB T2TT}
Finally, we evaluate KUTED for text-to-text translation (T2TT). The NLLB 1.3B model is fine-tuned with a learning rate of $1\times 10^{-4}$ using Adam, a batch size of 16, and $100$k iterations on two RTX~8000 GPUs. The fine-tuned system attains 16.72 BLEU on EN$\rightarrow$CKB and 27.93 BLEU on CKB$\rightarrow$EN on the KUTED test set (Table~\ref{tab:t2tt}).

\begin{table}[!ht]
\centering
\footnotesize
\begin{tabularx}{\columnwidth}{lYY}
\toprule
\textbf{System} & \textbf{BLEU} & \textbf{ChrF++} \\
\midrule
EN$\rightarrow$CKB & 16.72 & 46.75 \\
CKB$\rightarrow$EN & 27.93 & 49.73 \\
\bottomrule
\end{tabularx}
\caption{T2TT results on the KUTED test set.}
\label{tab:t2tt}
\end{table}

\section{Conclusion}
\label{sec:foot}
We introduce KUTED, an English$\rightarrow$Central Kurdish speech translation corpus, comprising 170 hours of English speech aligned with English transcripts and Kurdish translations. We address Central Kurdish orthographic standardization and propose a systematic procedure for normalizing and correcting text. We evaluate KUTED across end-to-end (E2E) S2TT, cascaded S2TT, and T2TT settings, demonstrating that fine-tuning pretrained models such as Seamless yields substantial gains, including a +3 BLEU improvement on the out-of-domain FLEURS benchmark. We further assess the dataset with a Transformer-based S2TT model trained from scratch and report bidirectional T2TT results (EN$\rightarrow$CKB and CKB$\rightarrow$EN), underscoring KUTED’s utility for both speech and text translation. Future work includes extending this methodology to other low-resource languages with existing TED translations and automating script standardization by training models to map noisy inputs to standardized forms.

\section{Copyright and ethical issues}
The French adaptation of the GDPR\footnote{\url{https://www.legifrance.gouv.fr/jorf/id/JORFTEXT000044362034}} allows automatic scrapping for scientific purposes.
However, in accordance with TED’s current copyright policy, the dataset itself cannot be shared publicly, neither the models. Therefore, only the list of TED Talks IDs used in this research will be made available to ensure the reproducibility of the results. Consequently, there is no legal issues with using children voices as no audio data will be shared \footnote{\url{https://huggingface.co/datasets/aranemini/kutedlist}}. 


\section{Acknowledgments}
This research was conducted at the LIUM (Laboratoire d'Informatique de l'Université du Mans) laboratory. This work was partially performed using HPC resources from GENCI–IDRIS (Grant AD011012527) and received funding from the DGA/AID RAPID COMMUTE project. The authors thank the TED Kurdish translator community, especially the Hiwa Foundation, for their pioneering work in translating TED Talks into Kurdish. We also acknowledge the Department of English Language at the Faculty of Education, Koya University, where the majority of these translations were completed. Additionally, we appreciate the assistance of Lavin Azwar Omar, Wafa Idrees Omar, and Shajwan Muhammad Kwekha in facilitating the evaluation of translation samples. Daban Q. Jaff extends his gratitude for the support of the Deutscher Akademischer Austauschdienst (DAAD) through a PhD research grant (Grant No. 57645448) for his doctoral studies at the University of Erfurt (Host: Language and Its Structure, Prof. Dr. Beate Hampe).

\bibliographystyle{lrec2026-natbib}
\bibliography{lrec2026-example}

\end{document}